\documentclass{article}
\usepackage{amsmath}
\usepackage[margin=1in]{geometry}
\usepackage{setspace}
\usepackage{booktabs}
\usepackage{amssymb}
\usepackage{appendix}
\usepackage{tikz}
\usepackage{pgfplots}
\usepackage{graphicx} 
\usepackage{listings}
\usepackage{caption}
\usepackage{subcaption}

\pgfplotsset{compat=1.18}

\DeclareMathOperator*{\argmin}{arg\,min}

\title{Reconstructing Item Characteristic Curves using Fine-Tuned Large Language Models}
\author{Christopher Ormerod\\
	Cambium Assessment\\
	christopher.ormerod@gmail.com}
\date{January 2026}

\begin{document}
	
	\maketitle

	\begin{abstract}
		Traditional methods for determining assessment item parameters, such as difficulty and discrimination, rely heavily on expensive field testing to collect student performance data for Item Response Theory (IRT) calibration. This study introduces a novel approach that implicitly models these psychometric properties by fine-tuning Large Language Models (LLMs) to simulate student responses across a spectrum of latent abilities. Leveraging the Qwen-3 dense model series and Low-Rank Adaptation (LoRA), we train models to generate responses to multiple choice questions conditioned on discrete ability descriptors. We reconstruct the probability of a correct response as a function of student ability, effectively generating synthetic Item Characteristic Curves (ICCs) to estimate IRT parameters. Evaluation on a dataset of Grade 6 English Language Arts (ELA) items and the BEA 2024 Shared Task dataset demonstrates that this method competes with or outperforms baseline approaches. This simulation-based technique seems particularly effective at modeling item discrimination.
	\end{abstract}
	
	\vspace{1cm}
	
	
	\section{Introduction}
	
	Item Difficulty Modeling (IDM) employs statistical techniques to describe and predict item difficulty as defined by Item Response Theory (IRT) \cite{hambleton_fundamentals_1991}. Determining difficulty has traditionally required curating student performance data to identify the best-fitting model for the probability of a correct response as a function of student ability \cite{rasch_probabilistic_1993}. Due to the high cost of this process, researchers have long sought to model difficulty as a function of textual features \cite{peters_text-based_2025}. Most approaches involve direct modeling, where the item difficulty parameters are modeled as explicit functions of features of the text, such as syntactic and readability measures. 
	
	Language models are probabilistic models that determine the likelihood of a sequence of words. Most LLMs are built on the transformer architecture \cite{vaswani_attention_2017} and are trained on large corpora to either predict the next token in a sequence \cite{radford_improving_2018} or predict masked tokens \cite{devlin_bert_2018}. When fine-tuned, these networks demonstrate remarkable abilities to understand, summarize, and generate text \cite{wang_glue_2019}. In educational applications, they have successfully performed automated scoring \cite{rodriguez_language_2019, sung_pre-training_2019}, the annotation of argumentative elements \cite{ormerod_argumentation_2023}, automatic item generation \cite{runge_generative_2024}, and IDM \cite{li_item_2025, yaneva_findings_2024}.
	
	Modern Generative LLMs have scaled the transformer architecture by an order of magnitude in terms of parameters \cite{openai_gpt-4_2023}. These models are trained to complete a wide range of tasks, and this scaling has given rise to emergent abilities \cite{wei_emergent_2022}. They excel in coding, mathematical problem-solving, and translation. While much educational research focuses on large foundational models \cite{openai_gpt-4_2023}, there are compelling reasons to consider open-source alternatives \cite{abdin_phi-3_2024, aimeta_llama_2024, yang_qwen2_2024}. Beyond offering privacy and security \cite{bulut_rise_2024}, parameter-efficient fine-tuning (PEFT) techniques \cite{xu_parameter-efficient_2023} allow researchers to control model outputs according to specific requirements in ways that standard prompt-tuning techniques have proven to be difficult \cite{giray_prompt_2023, singhvi_dspy_2024}. The core idea underlying this article's framing---that a fine-tuned language model can stand in for a population of examinees at a given ability level, so that item parameters can be recovered from the model's simulated responses rather than from field-collected data---was first articulated by Maeda, who fine-tuned transformer-based models according to the 2PL IRT model to generate synthetic multiple-choice responses for field-testing English grammar items \cite{maeda_field-testing_2023, maeda_field-testing_2024}. This article explores fine-tuning the Qwen-3 series models \cite{yang_qwen3_2025} using Low-Rank Adaptation (LoRA) \cite{hu_lora_2021} to implicitly perform IDM. We do this by training a model to simulate student response probabilities to multiple choice questions (MCQ) across varying ability levels. This approach aims to reconstruct the probability that a student correctly answers an unseen question as a function of their ability. We apply this method to a collection of English Language and Arts (ELA) items used in a state assessment program, which we call the ELA Dataset, and the Building Educational Applications (BEA) 2024 Shared Task dataset, which we call the BEA 2024 Shared Task. Given that these are very small datasets, the method employed demonstrates excellent performance compared with baselines. By approximating the specific errors students make as we vary ability, this method seems particularly well-suited to modeling discrimination. 
	
	While a version of this approach successfully modeled difficulty for items with free-form constructed responses \cite{scarlatos_smart_2025}, the way in which the models were trained in \cite{scarlatos_smart_2025} did not generalize naturally to MCQ items. This article is more directly aligned with the approach of Maeda \cite{maeda_field-testing_2023, maeda_field-testing_2024}, in which a transformer model (RoBERTa) is fine-tuned according to a target IRT model so that its output probabilities over answer choices reproduce a desired ICC; the present work extends this framing from an encoder-only model trained on a single ability parameter to a generative decoder LLM conditioned on discrete ability descriptors. For this article, the model is trained to reproduce token probabilities, where each token corresponds to a choice provided in an MCQ item. This work aligns with recent efforts to simulate LLM responses to Sentence Reading Efficiency tasks \cite{zelikman_generating_2023} and studies using the latent ability levels of various LLMs to estimate difficulty \cite{liu_leveraging_2025}.
	
	The remainder of this paper is organized as follows: Section \ref{sec:background} provides a comprehensive overview of the theoretical foundations, including IRT, the mechanics of LLMs, and the technical architecture of the Qwen-3 series. Section 3 details the methodology for item difficulty modeling, describing the discretization of ability level descriptors, the evaluation datasets, and the fine-tuning procedures used to simulate student responses. Section 4 presents experimental results from the ELA Dataset and BEA 2024 Shared Task datasets, followed by a discussion of the scaling laws observed across model sizes. Finally, the paper concludes with a summary of findings regarding the efficacy of using simulated ability levels to predict assessment difficulty.
	
	\section{Background}\label{sec:background}
	
	\subsection{Item Response Theory}
	
	For a dichotomous item, where a response $y$ is scored as either correct ($1$) or incorrect ($0$), the fundamental goal is to model the probability of a correct response as a function of a student's latent ability, denoted by $\theta$. This function is known as the Item Characteristic Curve (ICC). We approximate the ICC using the logistic function, specifically the Two-Parameter Logistic (2PL) and One-Parameter Logistic (1PL) models:
	\begin{equation}\label{eq:2Pl-dichotomous}
		\mathbb{P}(y = 1) = P_2(\theta; a, b) = \sigma\left(a (\theta - b)\right).
	\end{equation}
	where $\sigma(x) = 1/(1+e^{-x})$ is the sigmoid function, $a$ is the discrimination parameter, and $b$ is the difficulty parameter. The 1PL model is a constrained case of the 2PL model where $P_1(\theta; b) = P_2(\theta; 1, b)$.
	
	A field test collects responses from a sample of $N$ students, with assumed ability levels $\theta \sim \mathcal{N}(\mu, \varsigma^2)$ \footnote{We will use $\varsigma^2$ to denote variance instead of the standard $\sigma^2$ notation to distinguish the variance from the sigmoid function.}, across a set of items with parameters $a_{1,\ldots,M}$ and $b_{1,\ldots,M}$. The observed data consists of binary outcomes $y_{ij} \in \{0,1\}$ for all student-item pairs $(i,j)$ in the observed set $S$. Calibration involves estimating the item parameters and student abilities by minimizing the squared error between the theoretical probabilities and the observed responses:
	\begin{equation}
		\{\hat{\theta}_i\}, \{\hat{a}_j\}, \{\hat{b}_j\} = \argmin_{\theta, a, b} \sum_{(i,j) \in S} \left( P_2(\theta_i; a_j, b_j) - y_{ij} \right)^2. \label{eq:irt_optim}
	\end{equation}
	In this optimization, the student abilities $\theta_i$ may either be treated as free variables to be estimated or fixed to values derived from an external assessment context.
	
	In the case of multiple-choice items, where the observations can be any element of a finite set $V = \{v_1, \ldots, v_n\}$, the form of \eqref{eq:2Pl-dichotomous} naturally generalizes. We use the ansatz that the probability of each outcome is the result of the softmax operator applied to a collection of $n$ linear functions of $\theta$. We parameterize this as:
	\begin{equation}\left( \mathbb{P}(y_{ij} = v_1), \ldots, \mathbb{P}(y_{ij} = v_n) \right) = \mathrm{softmax}\left(a_1(\theta - b_1), \ldots, a_n (\theta - b_n) \right). 
		\label{eq:nominal}
	\end{equation}
	For $n=2$, this is equivalent to \eqref{eq:2Pl-dichotomous} for some $a$ and $b$. However, for arbitrary $n > 2$, the probability of a response being correct is not generally equivalent to an equation of the form \eqref{eq:2Pl-dichotomous}. This generalization is known as the Nominal Response Model (NRM).
	
	A special case of \eqref{eq:nominal} that remains equivalent to \eqref{eq:2Pl-dichotomous} occurs when we assume that if a student does not know the answer, the other options are chosen with equal probability. Without loss of generality, we may assume the correct answer is $v_1$. Since all other answers are equally probable, we can assume that $a_i = b_i = 0$ for $i \neq 1$. In this case, a simple calculation shows that:
	\[
	\mathbb{P}(y_{ij} = v_1) = \sigma\left( a_1\left(\theta - b_1\right) - \log(n-1) \right).
	\]
	This implies that the correspondence between the variables of \eqref{eq:2Pl-dichotomous} and \eqref{eq:nominal}, under the assumption that test-takers choose uniformly at random from the remaining options when incorrect, is given by:
	\begin{equation}\hat{a}_j = a_1, \quad \hat{b}_j = b_1 + \frac{\log(n-1)}{a_1} \label{eq:nrm_2pl_correspondence}
	\end{equation}
	where all other $a_i$ and $b_i$ values are $0$.
	
	\subsection{Language Models}
	
	At their fundamental level, generative LLMs function as autoregressive probabilistic models that estimate the conditional probability of a token given a specific context. The core task is to model the probability distribution of the next token, $w_t$, provided the sequence of all preceding tokens, $w_{1}, \ldots, w_{t-1}$:
	\begin{equation}
		P(w_t | w_1, w_2, ..., w_{t-1}; \Omega),
	\end{equation}
	where each $w_i$ represents a token from a finite fixed vocabulary $V = \{v_i\}$ and $\Omega$ represents the learned parameters (weights) of the neural network. A language model does not output a single prediction directly. Instead, the final layer produces a vector of logits, $z = z(\Omega; w)$, where the $i$-th element, $z_i$, is associated with the event $w_t = v_i$. The softmax function normalizes these logits into a valid probability distribution:
	\begin{equation}
		P(w_t = v_i) = \frac{e^{z_i}}{\sum_{j=1}^{|V|} e^{z_j}} \label{eq:llm_prob}
	\end{equation}
	This assigns a probability score between 0 and 1 to every possible next token, such that the sum of all probabilities equals 1. LLMs are trained using Maximum Likelihood Estimation (MLE). The goal is to maximize the probability assigned to the actual next token found in the training data by minimizing the Cross-Entropy Loss (Negative Log-Likelihood):
	\begin{equation}
		\mathcal{L} = -\sum_{t} \log P(w_t^{target} | w_{1:t-1}).
	\end{equation}
	In the Transformer architecture \cite{vaswani_attention_2017}, attention—specifically masked self-attention—is the mechanism that allows the model to ``understand" the context of a sentence by determining which previous tokens are relevant to the current prediction \cite{devlin_bert_2018, radford_improving_2018}. Instead of treating all preceding words equally or focusing only on the most recent one, attention assigns a dynamic ``relevance score" to every past token. This allows the model to construct a context vector that is a weighted mixture of the entire history. To process context, the model transforms every token in the sequence into three vectors:
	\begin{enumerate}
		\item {\bf Query} ($Q$): Represents the current token asking, ``What information do I need to predict the next word?".
		\item {\bf Key} ($K$): Represents a preceding token answering, ``Here is what I define/contain.".
		\item {\bf Value} ($V$): The actual content or meaning of that preceding token.
	\end{enumerate}
	If $X \in \mathbb{R}^{T\times d}$ is a layers input where $d$ is some hidden dimension of the model with $T$ being the number of tokens, then 
	\begin{equation}
		Q = XW_q + b_q, \hspace{1cm} K = X W_k + b_k, \hspace{1cm} V = XW_v + b_v. \label{eq:linear_qkv}
	\end{equation}
	where $b_q$, $b_k$, and $b_v$ are called the bias terms. Typically, $W_q, W_k, W_v \in \mathbb{R}^{d \times d_k}$ where $d_k = d/h$ where $h$ is the number of attention heads. In some models, such as Llama, these bias terms are zero \cite{aimeta_llama_2024}, however, in the Qwen model series, these terms are nonzero \cite{yang_qwen3_2025}. When the model tries to predict the next token (time step $t$), it compares the Query of the current position against the Keys of all previous positions.
	
	Once a model is pretrained to determine the next word, most modern LLMs undergo two additional phases of training. First the models are subjected to supervised fine-tuning followed by a refinement of the outputs, sometimes called safety-tuning \cite{aimeta_llama_2024}. The supervised fine-tuning is a training in which the model is tuned to complete a set of tasks. Each task is encoded into components of the following form:
	
	\begin{verbatim}
		system: {system text specifying how the assistant should behave}
		user: {specification of the task}
		assistant: {completion of the task}
		user: {subsequent follow-up task}
		assistant: {subsequent follow-up completion}
		...
		assistant: {last completion}
	\end{verbatim}
	This input can be parsed and presented as a back-and-forth chat in web-based user interfaces. While the web-interface has broken down the technical barriers to using LLMs and made it seem as if these models have personalities of their own, it is still fundamentally a model focused on calculating the probability of the next token.
	
	A feature of many modern LLMs is to also include chain-of-thought processes into the solution \cite{wei_chain--thought_2023}, which can be encoded in a similar manner as \verb|thinking|, however, we can also consider this to be a subcomponent of the \verb|assistant| component. The refinement process uses proximal policy optimization often associated with reinforcement learning \cite{schulman_proximal_2017}. The process involves pairs or collections of completions in the form above, where one of the completions is considered optimal with respect to some human preferences (safe and useful) or some model based on preferences \cite{ouyang_training_2022}.  
	
	\subsection{Qwen Models}
	
	As of late 2025, the Qwen 3 series (developed by Alibaba Cloud) represents a significant evolution in open-weights large language models, notable for introducing a highly granular range of model sizes and hybrid architectures \cite{yang_qwen3_2025}. The Qwen 3 family includes both dense and Mixture-of-Experts (MoE) architectures, designed to bridge the gap between lightweight edge deployment and massive high-performance computing. The dense Qwen 3 suite is particularly well-suited for studying the scaling laws of neural networks (how model performance changes as parameter count increases) since Qwen 3 offers a smooth continuum of sizes
	\[
	0.6B \rightarrow 1.7B \rightarrow 4B \rightarrow 8B \rightarrow 14B \rightarrow 32B
	\]
	The Llama models, which are perhaps the best open-source alternative at this time, have 1B, 3B, 8B, and 70B variants  \cite{aimeta_llama_2024}. In many instances, the hardware made available made experimentation with 70B and even 32B models difficult purely in terms of the combination of the size of the input and the model sizes. 
	
	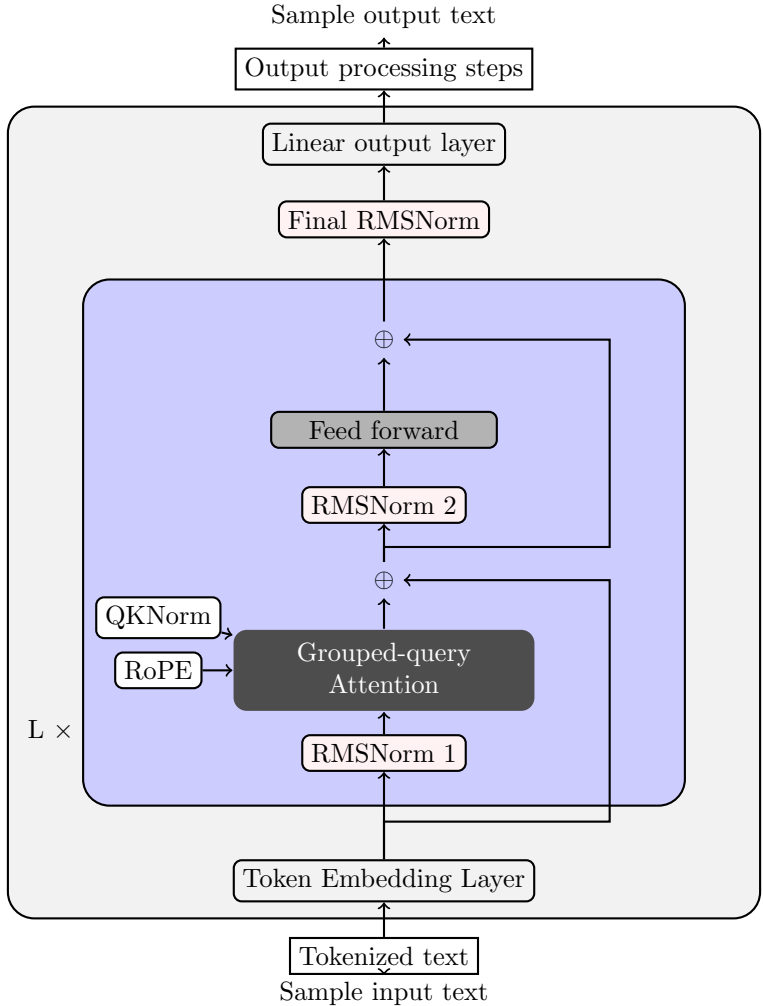
\begin{figure}[!ht]
		\centering
		\begin{tikzpicture}
			\draw[rounded corners = 10pt, fill=black!5, draw=black, thick] (-5,-4.5) rectangle (5,6.3); 
			
			\node (input) at (0,-5.5) {Sample input text};
			\node[draw, thick] (tokenized) at (0,-5) {Tokenized text};
			\node[rounded corners = 3pt, fill=black!5, draw=black, thick] (embed) at (0,-4) {Token Embedding Layer};
			
			\draw[rounded corners = 10pt, fill=blue!20, draw=black, thick] (-4,-3) rectangle (4,4); 
			
			\node[rounded corners = 3pt, fill=red!5, draw=black, thick] (rms1) at (0,-2.3) {RMSNorm 1};
			\node[rounded corners = 5pt, fill=black!70, text=white, thick, minimum width=4cm] (attn) at (0,-1.2) {\begin{tabular}{c}Grouped-query\\ Attention\end{tabular}};
			\node[rounded corners = 3pt, fill=white, draw=black, thick] (rope) at (-3,-1.2) {RoPE};
			\node[rounded corners = 3pt, fill=white, draw=black, thick] (qknorm) at (-3,-0.5) {QKNorm};
			
			\node (plus1) at (0,0) {$\oplus$};
			\node[rounded corners = 3pt, fill=red!5, draw=black, thick] (rms2) at (0,1) {RMSNorm 2};
			\node[rounded corners = 3pt, fill=black!30, draw=black, thick, minimum width=3cm] (ffn) at (0,2) {Feed forward};
			\node (plus2) at (0,3.2) {$\oplus$};
			
			\node[rounded corners = 3pt, fill=red!5, draw=black, thick] (finalrms) at (0,4.8) {Final RMSNorm};
			\node[rounded corners = 3pt, fill=black!5, draw=black, thick] (linear) at (0,5.8) {Linear output layer};
			\node[draw, thick] (proc) at (0,6.8) {Output processing steps};
			\node (output) at (0,7.5) {Sample output text};
			
			\draw[->, thick] (input) -- (tokenized);
			\draw[->, thick] (tokenized) -- (embed);
			\draw[->, thick] (embed) -- (rms1);
			\draw[->, thick] (rms1) -- (attn);
			\draw[->, thick] (attn) -- (plus1);
			\draw[->, thick] (plus1) -- (rms2);
			\draw[->, thick] (rms2) -- (ffn);
			\draw[->, thick] (ffn) -- (plus2);
			\draw[->, thick] (plus2) -- (finalrms);
			\draw[->, thick] (finalrms) -- (linear);
			\draw[->, thick] (linear) -- (proc);
			\draw[->, thick] (proc) -- (output);
			
			\draw[->, thick] (embed.north) -- ++(0,0.5) -- ++(3,0) |- (plus1.east);
			\draw[->, thick] (plus1.north) -- ++(0,0.2) -- ++(3,0) |- (plus2.east);
			
			\draw[->, thick] (rope) -- (attn);
			\draw[->, thick] (qknorm) -- (attn);
			
			\node at (-4, -2) [left] {L $\times$};
		\end{tikzpicture}
		\caption{Architectural Schematic of the Qwen3 Dense Model Series. This diagram illustrates the transformer block structure used across the Qwen3 lineup, highlighting the implementation of Root Mean Square Layer Normalization (RMSNorm) and Masked Grouped-query Attention. The architecture features a ``pre-norm" configuration where RMSNorm 1 and RMSNorm 2 precede the attention and feed-forward layers, respectively. Key technical enhancements include Rotary Positional Embeddings (RoPE) and QKNorm to improve training stability.}
		\label{fig:qwen_architecture}
	\end{figure}
	
	The architecture of these models is presented in Figure \ref{fig:qwen_architecture}. With this in mind, there are just a few parameters governing the architecture that determine the size of the model. The most important of these parameters are the number of layers, the hidden size of the model, the number of attention heads, and their groups. Table \ref{tab:qwen3_specs} presents the technical specifications of these models.
	
	\begin{table}[ht]
		\centering
		\begin{tabular}{lccccc}
			\toprule
			&&& \textbf{Hidden} & \textbf{Attention} & \textbf{Vocab} \\
			\textbf{Model} & \textbf{Parameters} & \textbf{Layers} & \textbf{Size} & \textbf{Heads} & \textbf{Size} \\
			\midrule
			Qwen3-1.7B & 1.72 B & 28 & 2,048 & 16 / 8 & 151,936 \\
			Qwen3-4B   & 4.02 B & 36 & 2,560 & 32 / 8 & 151,936 \\
			Qwen3-8B   & 8.19 B & 36 & 4,096 & 32 / 8 & 151,936 \\
			Qwen3-14B  & 14.77 B & 40 & 5,120 & 40 / 8 & 151,936 \\
			Qwen3-32B  & 32.76 B & 64 & 8,192 & 64 / 8 & 151,936 \\
			\bottomrule
		\end{tabular}
		\caption{Structural Key Characteristics of Qwen3 Dense Models.}
		\label{tab:qwen3_specs}
	\end{table}
	
	All models utilize a head dimension of 128, consistent with previous Qwen architectures \cite{yang_qwen2_2024}. The entire lineup uses Grouped Query Attention (GQA). For example, the 32B model has 64 query heads but only 8 key/value heads, significantly reducing KV cache memory. The 1.7B and 4B models support a native context of 32K tokens, while the 8B and 14B support 128K tokens natively. The relevant structural information of each of the models used in this study have been presented in Table \ref{tab:qwen3_specs}.
	
	\section{Method}
	
	The method in this article consists of defining three different probability values which provide three different discrete ICCs:
	\begin{enumerate}
		\item The observed probabilities: The proportion of students that an answer an item correctly given a collection ranges of $\theta$ values.
		\item The nominal response model probabilities: The values arising from fitting a model to the observed probabilities. 
		\item The large language model probabilities: The token probabilities associated with the correct answer when simulating responses from a particular ability level. 
	\end{enumerate}
	This section will detail how we arrive at each of these, and how we will calculate the item parameters from this. 
	
	\subsection{Ability Level Descriptors}
	
	The goal of this subsection will be to elucidate a framework where we are provided with labels instead of values of $\theta$. Instead of having $N$ students, we will be providing $N$ classes of students, each associated with a descriptor. This gives us $N$ labels, $L_1, \ldots, L_N$, where each label, $L_k$, is associated with bounds $(c_{k-1}, c_k)$. That is to say that a student is given label $L_k$ if $c_{k-1} < \theta \leq c_k$. By letting $c_0 = -\infty$ and $c_N = \infty$, then each $\theta \in \mathbb{R}$ is uniquely associated with a label, $L_k$. 
	
	These descriptors categorize distinct ability levels. Given that they serve as inputs for LLM possessing latent semantic knowledge, it is desirable to maintain a direct correspondence with the competency levels they define. We prompted a language model to provide appropriate descriptors and bounding values for use in this study. These descriptors and their bounding values are provided in Table \ref{tab:ability_descriptors}. 
	\begin{table}[!ht]
		\centering
		\begin{tabular}{l l | l l} \toprule
			Descriptor ($L_k$) & $(c_{k-1},c_k)$ & Descriptor ($L_k$) & $(c_{k-1},c_k)$ \\ \midrule
			Critical & $(\infty, -3)$ & Satisfactory & $(0,0.3)$\\
			Severely Limited & $(-3,-2.7)$ & Competent & $(0.3,0.6)$ \\
			Deficient & $(-2.7, -2.4)$ & Proficient & $(0.6,0.9)$\\
			Inadequate & $(-2.4, -2.1)$ & Accomplished & $(0.9,1.2)$\\
			Minimal & $(-2.1, -1.8)$ & Advanced & $(1.2, 1.5)$\\
			Emerging & $(-1.5, -1.2)$ & Superior & $(1.5, 1.8)$\\
			Developing & $(-1.2, -0.9)$ & Exceptional & $(1.8,2.1)$\\
			Approaching Basic & $(-0.9,-0.6)$ & Outstanding & $(2.1, 2.4)$\\
			Basic & $(-0.6,-0.3)$ & Distinguished & $(2.4,2.7)$\\
			Functional & $(-0.3,0)$ & Exemplary & $(2.7,\infty)$\\ \bottomrule
		\end{tabular}
		\caption{The descriptors used in this project and their respective bounds.} \label{tab:ability_descriptors}
	\end{table}
	
	While the semantic distinction between labels associated with proximate $\theta$ values, such as ``Exceptional" versus ``Outstanding", may be negligible, the semantic distinctions become clearer as the intervals they represent grow further apart. In any case, we will fine-tune response probabilities to reinforce the adherence between specific descriptors and their targeted ability intervals. In our discrete setting, the discrete ICC for item $j$ is given by the $N$ probabilities, $(P_{j1}, \ldots , P_{jn})$, where each $P_{jk}$ is the probability that a student in the class of students with label $L_k$ will provide a correct answer
	\begin{equation}
		P_{jk} = \mathbb{P}\left(c_{k-1} < \theta_i \leq c_k\, |\, y_{ij} = 1 \right).
	\end{equation}
	
	Given the discrete ICC, in order to map back to continuous values of the difficulty and discrimination parameters, we map the label $L_k$ to the expected $\theta$ values on the class of $\theta$ values with the label $L_k$, denoted $\overline{\theta}_k$. To derive these, we use the standard assumption that $\theta \sim \mathcal{N}(\mu,\varsigma^2)$. Using the standard notation for the Probability Distribution Function (PDF) and Cumulative Distribution Function (CDF)
	\begin{eqnarray*}
		&\phi(x) = \frac{1}{\sqrt{2\pi}} \exp \left(\frac{x^2}{2}\right), \hspace{2cm} & \Phi(x) = \int_0^x \phi(z) \mathrm{d}z, \\
		&f(x;\mu,\varsigma) = \frac{1}{\varsigma}\phi\left( \frac{x-\mu}{\varsigma} \right), & F(x;\mu, \varsigma) = \Phi\left( \frac{x-\mu}{\varsigma} \right),
	\end{eqnarray*}
	we express the expected values, $\overline{\theta}_k$, by
	\[
	\overline{\theta}_k = \mathbb{E}(\theta\, |\, c_{k-1} < \theta \leq c_k) = \mu + \varsigma^2 \left(\frac{f(c_{k-1};\mu, \varsigma) - f(c_k;\mu, \varsigma)}{F(c_k;\mu,\varsigma) - F(c_{k-1};\mu, \varsigma)} \right).
	\]
	When minimizing \eqref{eq:irt_optim}, we also need to take into account how often terms of a given $\theta$ will appear in the sum. We take this into account by defining weights, $\omega_k$, based on probability of the label $L_k$ being applied:
	\begin{equation}
		\omega_k = F(c_k; \mu, \varsigma) - F(c_{k-1}; \mu, \varsigma) \label{eq:weights}
	\end{equation}
	We may recover reasonable approximations of the difficulty parameters by considering the minimization problem
	\begin{equation}
		a_j, b_j = \argmin_{a_j, b_j} \sum_{k} \omega_k \left(P_{jk} - \sigma\left( a_j(\overline{\theta}_k - b_j) \right) \right)^2. \label{eq:dicc_diff} 
	\end{equation}
	These values can be obtained by any number of minimization techniques. We typically used Broyden- Fletcher-Goldfarb-Shanno (BFGS) algorithm in this study. This provides a method of determining a reasonable approximation of difficulty from a discrete ICC. 
	
	We are now required to give approximations of the $P_{jk}$ values from empirical observations. Given a collection of students, each with a specific $\theta$ value, one method of approximating the probability $P_{jk}$ would be to take the ratio of correct to incorrect answers to item $j$ for all students in which label $L_k$ has been applied. 
	\begin{eqnarray*}
		C_{ijk} &=& \textrm{\# of students with label $L_k$ responding with $v_i$ to item $j$,}\\
		C_{jk} &=& \sum_i C_{ijk} = \textrm{\# of students with label $L_k$ responding to item $j$,}\\
		C_{j} &=& \sum_k C_{jk} = \textrm{\# of students responding to item $j$.}
	\end{eqnarray*}
	Without loss of generality, we assume the correct answer is given by $v_1$. This creates the first version of the ICC for item $j$, provided by
	\begin{align}
		(P_{j1}, \ldots , P_{jn}) = \left( \frac{C_{1j1}}{C_{j1}}, \ldots, \frac{C_{1j1}}{C_{j1}} \right) 
	\end{align}
	This may seem appealing given its simplicity, but this quantity can be a bad estimate with very few observations associated with a particular label or even ill-defined when there no student responses for some labels.
	
	Rather than use the empirical observations directly to provide the values of $P_{jk}$, we use the empirical observations to fit a NRM. Given an item, $j$, a label $k$, and a response $v_i$, we use $\left(\overline{\theta}_k, C_{ijk}/C_{jk}\right)$ as a datapoint for fitting our NRM. However, we must also appropriately weight each of these datapoints by the class probability on that item, provided by $C_{jk}/C_k$. This means that the parameters of our NRM for item $j$ are defined by
	\begin{equation}
		\{ a_{ij}\}, \{b_{ij}\} = \argmin_{a_{ij},b_{ij}} \sum_{i,k} \frac{C_{jk}}{C_j} \left( \frac{C_{ijk}}{C_{jk}} - \frac{\exp( a_{ij}(\overline{\theta}_k - b_{ij})}{\sum_g \exp( a_{gj}(\overline{\theta}_k - b_{gj})} \right)^2 \label{eq:smoother}
	\end{equation}
	Once these values are defined, this provides us with a model for the probability that a student of ability level $k$ answers with $v_i$ to item $j$:
	\begin{equation}
		P_{jk} = \frac{\exp( a_{ij}(\overline{\theta}_k - b_{ij})}{\sum_g \exp( a_{gj}(\overline{\theta}_k - b_{gj})} \label{eq:final_probs}
	\end{equation}
	If we again assume that $v_1$ is the correct answer, then the associated discrete ICC is provided by
	\[
	(P_{j1}, \ldots , P_{jn}) = \left( \frac{\exp( a_{1j}(\overline{\theta}_n - b_{1j})}{\sum_g \exp( a_{gj}(\overline{\theta}_1 - b_{gj})}, \ldots, \frac{\exp( a_{1j}(\overline{\theta}_n - b_{1j})}{\sum_g \exp( a_{gj}(\overline{\theta}_1 - b_{gj})} \right) 
	\]
	By minimizing \eqref{eq:smoother}, we expect that $P_{jk} \approx C_{ijk}/C_{jk}$. Furthermore, each $P_{jk}$ is informed by the entire curve rather than one observation, possibly making this approximation a more accurate representation of the probability associated with each label than $C_{ijk}/C_{jk}$. The weighting factor of $C_{jk}/C_j$ ensures that cases with few observations do not have a large effect on the calculation of $\{a_{ij}\}$ and $\{b_{ij}\}$. Perhaps the most useful way to understand this formulation is that \eqref{eq:final_probs} represents a smooth version of the empirical observed probabilities. 
	
	\subsection{Fine-tuning}
	
	The fundamental task of any generative LLM is to provide a function that determines the probability distribution of the next word/token based on the previous tokens \cite{radford_improving_2018}. Modern applications of generative LLMs primarily focus on the iterative application of this function to generate content; hence, most supervised and fine-tuning techniques have concentrated on manipulating the LLM weights in order to produce particular content. This paper takes a fundamentally different approach in that we are more concerned with the distribution of tokens than the actual token being produced.
	
	Given the size of these models, and the limits of our resources, it is still not possible to fine-tune the larger models. Due to the precision of the task, we stored the variables in 16-bit floating-point arithmetic and used low-rank adapters \cite{hu_lora_2021}. That is to say that instead of fine-tuning the model weights directly, we select a collection of linear layers and augment them by the addition of an additive factor. This means that applying LoRA substitutes the linear operator $L$ with $\tilde{L}$
	\begin{equation}
		L(x) = Mx + b \longrightarrow \tilde{L}(x) = (M + \Delta)x + b, \hspace{2cm} \Delta = BA,
	\end{equation}
	where $M$ is frozen in the optimization. We can generally replace any of the linear layers within the transformer architecture, however, it is more prudent to select a collection of ``target-layers" \cite{dettmers_qlora_2023}. The tried-and-true method has been to target the transformations of the input that define the keys, queries, and values \cite{ormerod_automated_2024}. In the context of Table \ref{tab:qwen3_specs}, the matrix $BA$ is a square matrix with rows and columns equal to the hidden size of the model. What makes this a parameter-efficient method is that $B$ possesses only $r$ columns and $A$ possesses only $r$ rows. Consequently, $\Delta$ is a matrix of low-rank. In our experiments, this rank is chosen to be $64$. 
	
	Once the model was augmented using LoRA, fine-tuning of the models was performed in two stages; supervised fine-tuning and distribution correction. During supervised fine-tuning, we take prompts specified in the form presented in Appendix \ref{sec:prompting}. We sampled a single answer for each item for each ability level randomly from the distribution of responses in a single epoch. This process fine-tunes the model to the production of output in an expected form. This fine-tuning used the cross-entropy loss, and the main objective of fine-tuning is to change the model outputs to produce an expected answer in the correct spot. We trained the model in this way for only one epoch using a version of the Adam optimizer \cite{loshchilov_decoupled_2019} with a learning rate of $5\times 10^{-5}$ and a batch size of one to accommodate the large input size with the usual linear learning rate scheduler and gradient clipping. 
	
	In the second phase of training, we truncate the prompt specified in Appendix \ref{sec:prompting} up to the point at which the answer is presented. Given the first phase of training, we are now looking to equate the output probabilities defined by the language model, given by \eqref{eq:llm_prob} over the set of tokens corresponding to an option, with the probabilities of each option defined by the nominal model, provided by \eqref{eq:nominal}. That is to say, we seek to equate
	\begin{equation}
		\left( \frac{e^{z_1}}{\sum_{j=1}^{n} e^{z_j}}, \ldots \frac{e^{z_n}}{\sum_{j=1}^{n} e^{z_j}} \right) \approx \mathrm{softmax} (a_1(\theta - b_1), \ldots , a_n(\theta - b_n)), \label{eq:llm:vs:nrm}
	\end{equation}
	where the left hand side involves the logits, $z_1, \ldots, z_n$, from \eqref{eq:llm_prob} associated choices $v_1, \ldots, v_n$ from \eqref{eq:nominal}. While we could equate the logits directly, one of the problems with this method is that they are not well-defined as the addition of a constant to all logits results in equivalent probabilities. Our loss function, from the standpoint of optimizing the LoRA adapter weights is the square of the difference between the left and right hand side of \eqref{eq:llm:vs:nrm}. We use a similar optimizer \cite{loshchilov_decoupled_2019} with a smaller learning rate of $5\times 10^{-6}$ with the usual linear learning rate scheduler and gradient clipping for stability.
	
	Over the course of a single epoch, the model is exposed to each question 20 times, where 20 is the number of ability level descriptors. We used the development set in an early stopping mechanism that optimizes the weighted difference \eqref{eq:llm:vs:nrm} where the weights are defined by \eqref{eq:weights} for each ability level descriptor. One final detail is that we expect this process to regress the difficulty estimates to the mean. To compensate for this, we fit the ability level estimates on the development set to the true ability level values with a linear regression, optimizing the Mean Squared Error on the development set. This linear transformation was also applied to the test set.
	
	\subsection{Data}
	
	\subsubsection{English Language Arts Dataset}
	
	The data consists of a collection of 275 multiple-choice items and associated student responses administered in a state-assessment program for Grade 6 English Language and Arts Assessment. The student ability levels were calibrated in the context of a larger item pool. In this way, we consider the student ability-levels, $\theta_i$, constant and the variables to optimize in \eqref{eq:irt_optim} to be $\{\hat{a}_j, \hat{b}_j\}$. Once the $\theta$ are fixed, equation \eqref{eq:irt_optim} can be optimized independently to determine the difficulty and discrimination of each item. 
	
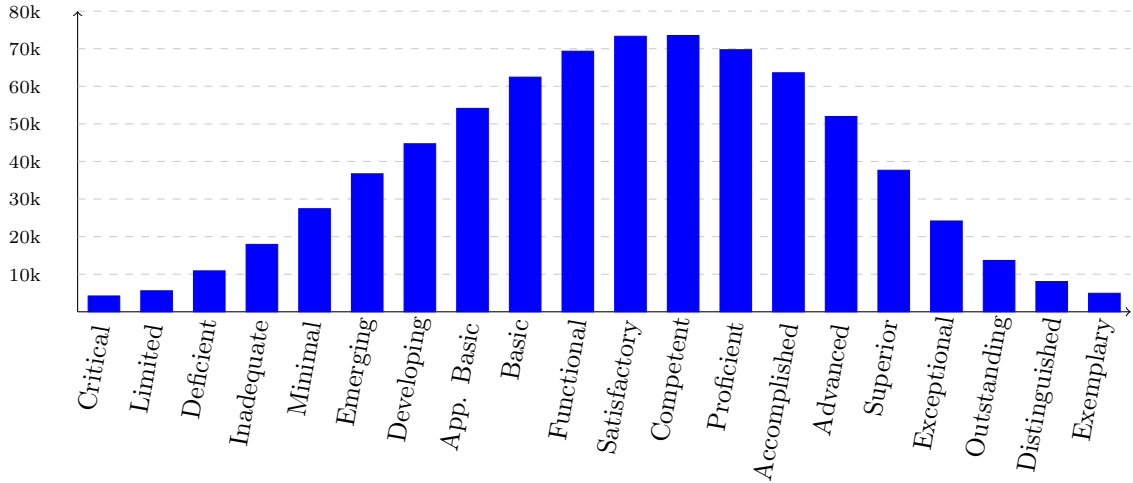
\begin{figure}
		\begin{tikzpicture}[yscale=0.05, xscale=0.7]
			\draw[->] (0,0) -- (20,0);
			\draw[->] (0,0) -- (0,80);
			\draw[dashed,draw=black!20] (0,10) -- (20,10);
			\draw[dashed,draw=black!20] (0,20) -- (20,20);
			\draw[dashed,draw=black!20] (0,30) -- (20,30);
			\draw[dashed,draw=black!20] (0,40) -- (20,40);
			\draw[dashed,draw=black!20] (0,50) -- (20,50);
			\draw[dashed,draw=black!20] (0,60) -- (20,60);
			\draw[dashed,draw=black!20] (0,70) -- (20,70);
			\draw[dashed,draw=black!20] (0,80) -- (20,80);
			\node at (-1,10) {\scriptsize 10k};
			\node at (-1,20) {\scriptsize 20k};
			\node at (-1,30) {\scriptsize 30k};
			\node at (-1,40) {\scriptsize 40k};
			\node at (-1,50) {\scriptsize 50k};
			\node at (-1,60) {\scriptsize 60k};
			\node at (-1,70) {\scriptsize 70k};
			\node at (-1,80) {\scriptsize 80k};
			\draw[draw=blue, fill=blue] (0.2,0) rectangle (.8,4.209);
			\draw[draw=blue, fill=blue] (1.2,0) rectangle (1.8,5.583);
			\draw[draw=blue, fill=blue] (2.2,0) rectangle (2.8,10.885);
			\draw[draw=blue, fill=blue] (3.2,0) rectangle (3.8,17.925);
			\draw[draw=blue, fill=blue] (4.2,0) rectangle (4.8,27.428);
			\draw[draw=blue, fill=blue] (5.2,0) rectangle (5.8,36.717);
			\draw[draw=blue, fill=blue] (6.2,0) rectangle (6.8,44.711);
			\draw[draw=blue, fill=blue] (7.2,0) rectangle (7.8,54.117);
			\draw[draw=blue, fill=blue] (8.2,0) rectangle (8.8,62.438);
			\draw[draw=blue, fill=blue] (9.2,0) rectangle (9.8,69.341);
			\draw[draw=blue, fill=blue] (10.2,0) rectangle (10.8,73.307);
			\draw[draw=blue, fill=blue] (11.2,0) rectangle (11.8,73.519);
			\draw[draw=blue, fill=blue] (12.2,0) rectangle (12.8,69.731);
			\draw[draw=blue, fill=blue] (13.2,0) rectangle (13.8,63.601);
			\draw[draw=blue, fill=blue] (14.2,0) rectangle (14.8,51.954);
			\draw[draw=blue, fill=blue] (15.2,0) rectangle (15.8,37.662);
			\draw[draw=blue, fill=blue] (16.2,0) rectangle (16.8,24.159);
			\draw[draw=blue, fill=blue] (17.2,0) rectangle (17.8,13.658);
			\draw[draw=blue, fill=blue] (18.2,0) rectangle (18.8,8.046);
			\draw[draw=blue, fill=blue] (19.2,0) rectangle (19.8,4.932);
			\node at (0.3,-15) {\rotatebox[]{80}{Critical}};
			\node at (1.3,-15) {\rotatebox[]{80}{Limited}};
			\node at (2.3,-15) {\rotatebox[]{80}{Deficient}};
			\node at (3.3,-19) {\rotatebox[]{80}{Inadequate}};
			\node at (4.3,-15) {\rotatebox[]{80}{Minimal}};
			\node at (5.3,-16) {\rotatebox[]{80}{Emerging}};
			\node at (6.3,-18) {\rotatebox[]{80}{Developing}};
			\node at (7.3,-20) {\rotatebox[]{80}{App. Basic}};
			\node at (8.3,-10) {\rotatebox[]{80}{Basic}};
			\node at (9.3,-18) {\rotatebox[]{80}{Functional}};
			\node at (10.3,-20) {\rotatebox[]{80}{Satisfactory}};
			\node at (11.3,-18) {\rotatebox[]{80}{Competent}};
			\node at (12.3,-17) {\rotatebox[]{80}{Proficient}};
			\node at (13.3,-24) {\rotatebox[]{80}{Accomplished}};
			\node at (14.3,-18) {\rotatebox[]{80}{Advanced}};
			\node at (15.3,-15) {\rotatebox[]{80}{Superior}};
			\node at (16.3,-19) {\rotatebox[]{80}{Exceptional}};
			\node at (17.3,-20) {\rotatebox[]{80}{Outstanding}};
			\node at (18.3,-22) {\rotatebox[]{80}{Distinguished}};
			\node at (19.3,-18) {\rotatebox[]{80}{Exemplary}};
		\end{tikzpicture}
		\caption{Distribution of Ability-Level Descriptors. The bar chart illustrates the frequency of student responses categorized by the 20 discrete descriptors, ranging from ``Critical" to ``Exemplary". The resulting distribution approximates a normal curve, consistent with the calibrated student ability parameters $\theta \sim \mathcal{N}(0.13, 1.15)$ observed in the Grade 6 English Language and Arts assessment data.}\label{fig:ability_dist}
	\end{figure}

	The dataset consisted of 754,000 student responses to 275 items, with item response frequencies ranging from 600 to 8,000 depending on the item. Each item has four options to choose from. The distribution of student ability parameters was approximately normal, where $\theta \sim \mathcal{N}(0.13, 1.15)$. After applying the ability level descriptors, the frequency of each ability-level descriptor, shown in Figure \ref{fig:ability_dist} approximates a normal curve. The distribution of $\theta$ values at the item level for each item is close to this distribution, which means that some items are expected to have, and have few or even no responses associated with a particular ability-level descriptor. 
	
	\subsubsection{BEA 2024 Shared Task}
	
	The BEA 2024 Shared Task, formally titled ``Automated Prediction of Item Difficulty and Item Response Time", was a competition organized by the National Board of Medical Examiners (NBME) to determine how effectively we can analyze the difficulty of exam questions without relying on live pre-testing. Held at the 19th Workshop on Innovative Use of NLP for Building Educational Applications (BEA 2024), the task focused on two main objectives:
	\begin{enumerate}
		\item{{\bf Item Difficulty Prediction:} Participants built models to predict the difficulty of a question.}
		\item{{\bf Item Response Time Prediction:} Participants predicted the average time (in seconds) test-takers would need to answer the question.}
	\end{enumerate}
	The data presented included the difficulty, as defined by the 1PL model. Since we do not have responses or probabilities directly to estimate the other response probabilities, we rely on the approximation that incorrect choices are uniformly random. This allows us to use the correspondence provided by \eqref{eq:nrm_2pl_correspondence}. 
	
	In terms of characteristics, the number of items released for training was 466, which was randomly split into a train set and a development set, while the test set consisted of 201 items. The distribution of the difficulties in each set was approximately normal with means ($\mu$) and standard deviation ($\varsigma$) provided in Table \ref{tab:bea:distribution}. 
	One aspect of this dataset that is worth mentioning is that the number of options in the test set seems to have been taken from a different distribution from those chosen for training. The test set has a much higher average number of options, which could make the task of predicting difficulty on this particular test set more difficult than one that more closely resembles the training set. The work of Bulut et. al showed exceptional performance when a subset of the training set was used as a test set \cite{bulut_item_2024}. While the task of predicting the time taken was also a part of the task, modeling timing is not the focus of this paper. 
	
	\begin{table}[!ht]
		\centering
		\begin{tabular}{l r | r r | r r r r r r r r} \toprule
			&& \multicolumn{2}{|c}{Difficulty} & \multicolumn{7}{c}{Number of Options}\\
			& N & $\mu$ & $\varsigma$ & 4 & 5 & 6 & 7 & 8 & 9 & 10\\ \midrule
			Train & 419 & 0.484 & 0.309 & 25 & 333 & 45 & 6 & 7 & 2 & 1\\
			Dev & 47 & 0.472 & 0.288 & 7 & 33 & 6 & 0 & 0 &1 & 0 \\
			Test & 201 & 0.500 & 0.310 & 0 & 0 & 15 & 159 & 20 & 2 & 10\\ \midrule
			Total & 667 & 0.488 & 0.309 &32 & 366 & 66 & 165 & 27 & 5 & 11 \\ \bottomrule
		\end{tabular}
		\caption{The distributional characteristics of the difficulty parameters for the BEA 2024 Shared Task.}\label{tab:bea:distribution}
	\end{table}
	
	\section{Results}
	
	To verify that the LLM is effectively simulating responses from students of varying ability levels, we compared three different discrete Item Characteristic Curves (ICCs). These include the observed empirical probabilities, the smoothed calibration probabilities derived from the Nominal Response Model (NRM), and the final LLM-generated probabilities.
	
	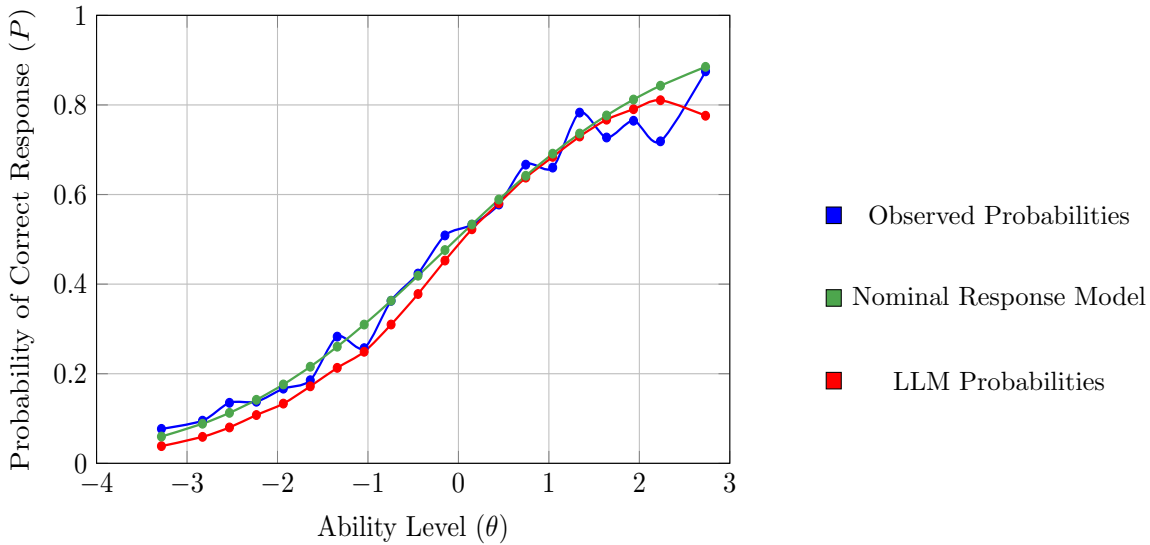
\begin{figure}[!ht]
		\begin{tikzpicture}[yscale=1.1]
\begin{axis}[
    xlabel={Ability Level ($\theta$)},
    ylabel={Probability of Correct Response ($P$)},
    xmin=-4, xmax=3,
    ymin=0, ymax=1,
    grid=major,
    width=10cm,
    height=7cm,
    scatter/classes={
        a={mark=*,blue}
    }
]

\addplot[
    smooth,
    blue,
    thick,
] coordinates {
    (-3.2831, 0.0769) (-2.8289, 0.0952) (-2.5311, 0.1351) (-2.2333, 0.1375) 
    (-1.9355, 0.1667) (-1.6377, 0.1855) (-1.3399, 0.2828) (-1.0422, 0.2571) 
    (-0.7444, 0.3627) (-0.4466, 0.4236) (-0.1489, 0.5087) (0.1489, 0.5324) 
    (0.4466, 0.5777) (0.7444, 0.6667) (1.0422, 0.6601) (1.3399, 0.7829) 
    (1.6377, 0.7273) (1.9355, 0.7647) (2.2333, 0.7188) (2.7319, 0.875)
};

\addplot[
    only marks,
    mark=*,
    blue,
    mark size=1.5pt
] coordinates {
    (-3.2831, 0.0769) (-2.8289, 0.0952) (-2.5311, 0.1351) (-2.2333, 0.1375) 
    (-1.9355, 0.1667) (-1.6377, 0.1855) (-1.3399, 0.2828) (-1.0422, 0.2571) 
    (-0.7444, 0.3627) (-0.4466, 0.4236) (-0.1489, 0.5087) (0.1489, 0.5324) 
    (0.4466, 0.5777) (0.7444, 0.6667) (1.0422, 0.6601) (1.3399, 0.7829) 
    (1.6377, 0.7273) (1.9355, 0.7647) (2.2333, 0.7188) (2.7319, 0.875)
};

\addplot[
    smooth,
    red,
    thick,
] coordinates {
    (-3.28309865493044, 0.038421630859375) (-2.8289428670255665, 0.05908203125) (-2.53111473282656, 0.080078125) (-2.233301264424591, 0.1077880859375) (-1.9355008677526035, 0.13330078125) (-1.6377119007555698, 0.171875) (-1.3399326789621737, 0.213134765625) (-1.042161481399394, 0.2490234375) (-0.744396556805868, 0.309814453125)  (-0.44663613009571596, 0.3779296875) (-0.14887840902111757, 0.45263671875) (0.14887840902111762, 0.5224609375)  (0.4466361300957155, 0.58056640625) (0.744396556805869, 0.6376953125) (1.0421614813993934, 0.68408203125) (1.3399326789621742, 0.7294921875) (1.637711900755568, 0.76708984375) (1.9355008677526082, 0.79052734375) (2.2333012644245915, 0.810546875) (2.7318611959539716, 0.77587890625)
};

\addplot[
    only marks,
    mark=*,
    red,
    mark size=1.5pt
] coordinates {
    (-3.28309865493044, 0.038421630859375) (-2.8289428670255665, 0.05908203125) (-2.53111473282656, 0.080078125) (-2.233301264424591, 0.1077880859375) (-1.9355008677526035, 0.13330078125) (-1.6377119007555698, 0.171875) (-1.3399326789621737, 0.213134765625) (-1.042161481399394, 0.2490234375) (-0.744396556805868, 0.309814453125)  (-0.44663613009571596, 0.3779296875) (-0.14887840902111757, 0.45263671875) (0.14887840902111762, 0.5224609375)  (0.4466361300957155, 0.58056640625) (0.744396556805869, 0.6376953125) (1.0421614813993934, 0.68408203125) (1.3399326789621742, 0.7294921875) (1.637711900755568, 0.76708984375) (1.9355008677526082, 0.79052734375) (2.2333012644245915, 0.810546875) (2.7318611959539716, 0.77587890625)
};

\addplot[
    smooth,
    green!50!black!70,
    thick,
] coordinates {
(-3.28309865493044, 0.0597399442256803) (-2.8289428670255665, 0.08840718484621282) (-2.53111473282656, 0.1126467333072167) (-2.233301264424591, 0.14175183273631695) (-1.9355008677526035, 0.17604538327710428) (-1.6377119007555698, 0.21563836964090297) (-1.3399326789621737, 0.26036396688929575) (-1.042161481399394, 0.3097315903462093) (-0.744396556805868, 0.36291630514034745)(-0.44663613009571596, 0.41879465257647497) (-0.14887840902111757, 0.4760283137824396) (0.14887840902111762, 0.5331845581134727) (0.4466361300957155, 0.5888717396967326) (0.744396556805869, 0.6418637973304356) (1.0421614813993934, 0.6911916673350877) (1.3399326789621742, 0.7361897835278237) (1.637711900755568, 0.7764978923864724) (1.9355008677526082, 0.8120278035941053) (2.2333012644245915, 0.8429091176202792) (2.7318611959539716, 0.8850312106487703)
};

\addplot[
    only marks,
    mark=*,
    green!50!black!70,
    mark size=1.5pt
] coordinates {
(-3.28309865493044, 0.0597399442256803) (-2.8289428670255665, 0.08840718484621282) (-2.53111473282656, 0.1126467333072167) (-2.233301264424591, 0.14175183273631695) (-1.9355008677526035, 0.17604538327710428) (-1.6377119007555698, 0.21563836964090297) (-1.3399326789621737, 0.26036396688929575) (-1.042161481399394, 0.3097315903462093) (-0.744396556805868, 0.36291630514034745)(-0.44663613009571596, 0.41879465257647497) (-0.14887840902111757, 0.4760283137824396) (0.14887840902111762, 0.5331845581134727) (0.4466361300957155, 0.5888717396967326) (0.744396556805869, 0.6418637973304356) (1.0421614813993934, 0.6911916673350877) (1.3399326789621742, 0.7361897835278237) (1.637711900755568, 0.7764978923864724) (1.9355008677526082, 0.8120278035941053) (2.2333012644245915, 0.8429091176202792) (2.7318611959539716, 0.8850312106487703)
};
\end{axis}
\node at (12,1) {LLM Probabilities};
\node at (12,2) {Nominal Response Model};
\node at (12,3) {Observed Probabilities};
\draw[fill=blue] (9.7,2.9) rectangle (9.9,3.1);
\draw[fill=green!50!black!70] (9.7,1.9) rectangle (9.9,2.1);
\draw[fill=red] (9.7,0.9) rectangle (9.9,1.1);

\end{tikzpicture}
		\caption{Comparison of Item Characteristic Curves (ICCs). This plot illustrates the relationship between student ability levels ($\theta$) and the probability of a correct response ($P$) for a sample item. The blue curve represents the Observed Probabilities from empirical student data, while the green curve depicts the Calibration Probabilities derived from fitting the Nominal Response Model (NRM). The red curve shows the LLM Probabilities, demonstrating the language model's ability to simulate responses that adhere to specific ability level descriptors after supervised fine-tuning and distribution correction.}\label{fig:comparisonplot}
	\end{figure}
	
	As shown in the comparison plot in Figure \ref{fig:comparisonplot}, the LLM-generated response probabilities closely track the calibration curve, demonstrating the model's adherence to the discrete ability level descriptors.
	
	\subsection{English Language Arts Dataset}
	
	The ELA dataset, consisting of 754,000 student responses across 275 items, was used to evaluate the model's ability to approximate IRT parameters. Calibration of the student ability levels ($\theta$) revealed a distribution approximately following $\mathcal{N}(0.13, 1.15)$. When applying the discrete ability level descriptors, the frequency of student responses followed a normal curve, ranging from ``Critical" to ``Exemplary". Our baseline consists of a fine-tuned version of ModernBERT \cite{warner_smarter_2024}, in combination with a feature-based model.

		\begin{table}[h]
			\centering
			\begin{tabular}{lcccccc}
				\toprule
				& \multicolumn{2}{c}{$b$ (1PL)} & \multicolumn{2}{c}{$a$ (2PL)} & \multicolumn{2}{c}{$b$ (2PL)} \\
				\cmidrule(lr){2-3} \cmidrule(lr){4-5} \cmidrule(lr){6-7}
				Model & Pearson & RMSE & Pearson & RMSE & Pearson & RMSE \\
				\midrule
				Qwen-1.7B & 0.409 & 0.766 & 0.238 & 0.191 & 0.410 & 1.041 \\
				Qwen-4B   & 0.404 & 0.764 & 0.332 & 0.186 & 0.391 & 1.008 \\
				Qwen-8B   & 0.408 & 0.750 & 0.336 & 0.191 & 0.429 & 1.014 \\
				Qwen-14B  & 0.503 & 0.721 & 0.446 & 0.169 & 0.485 & 0.936 \\
				ModernBERT & 0.239 & 0.868 & 0.061 & 0.231 & 0.132 & 1.251 \\
				Features  & 0.160 & 0.827  & 0.194 & 0.200 & 0.040 & 1.098 \\ 
				\bottomrule
			\end{tabular}
			\caption{This table presents the Pearson correlation coefficients and Root Mean Squared Error (RMSE) across five folds for the Qwen3 model series. The metrics are categorized by the modeling regime: the One-Parameter Logistic (1PL) model for difficulty ($b$), and the Two-Parameter Logistic (2PL) model for both discrimination ($a$) and difficulty ($b$). Average values are provided for each model to illustrate performance scaling from the 1.7B to the 14B parameter variants.\label{tab:res:smarter}}
		\end{table}
		
		While the overall correlations are modest, it is important to note the limited size of the item pool used. Notably, this approach yielded a relatively high correlation and low Root Mean Square Error (RMSE) for the discrimination parameter ($a$) compared to the difficulty parameter ($b$). This is a significant result, as many difficulty prediction methods fail to accurately model item discrimination, and such results are frequently omitted from similar studies. Across the ELA dataset, the Qwen-14B model demonstrated the strongest performance, achieving an average correlation of 0.503 for the 1PL difficulty parameter and 0.446 for 2PL discrimination.
		
		\subsection{BEA 2024 Shared Task}
		
		For the BEA 2024 Shared Task, models were evaluated on their ability to predict item difficulty ($b$) as defined by the 1PL model across a test set of 201 items. Because student response data was not directly available, the uniform randomness of incorrect choices was assumed to utilize the correspondence in equation (4).
		
		\begin{table}
			\centering
			\begin{tabular}{llll}
				\toprule
				model & Pearson Correlation & RMSE \\
				\midrule
				Dummy Regressor Baseline \cite{yaneva_findings_2024} & & 0.31\\ 
				ELECTRA \cite{yaneva_findings_2024} & & 0.299\\ 
				Ensemble \cite{li_item_2025} && 0.292 \\
				\midrule
				Qwen-1.7B & 0.157 & 0.317 \\
				Qwen-4B & 0.212 & 0.309 \\
				Qwen-8B & 0.381 & 0.288 \\
				Qwen-14B & 0.336 & 0.294 \\
				Qwen-32B & 0.365 & 0.297 \\
				\bottomrule
			\end{tabular}
			\caption{This table presents the average Pearson correlation coefficients and Root Mean Squared Error (RMSE) values using the Qwen3 model series for the BEA 2024 Shared Task.\label{tab:res:nbme}}
		\end{table}
		
		The predictive accuracy of the Qwen3 series was compared against standard baselines (Table \ref{tab:res:nbme}):
		\begin{itemize}
			\item{Qwen-8B achieved the highest Pearson Correlation of 0.381 and the lowest Root Mean Square Error (RMSE) of 0.288. This seems to be an improvement over many previous results \cite{yaneva_findings_2024, li_item_2025}.}
			\item{Qwen-14B followed closely with a correlation with RMSE of 0.294.}
			\item{Qwen-4B performed slightly better than the Dummy Regressor Baseline but lagged behind the larger models with an RMSE of 0.309.}
			\item{All Qwen models larger than 4B outperformed the ELECTRA baseline (RMSE 0.299) and the Dummy Regressor (RMSE 0.31).}
		\end{itemize}
		
		\section{Discussion}
		
		The findings of this study suggest that utilizing large language models to simulate varied student ability levels provides a promising avenue for item difficulty modeling, though several limitations and future research directions remain. This approach to modeling item difficulty remains in its infancy, and significant improvements can be anticipated as prompting techniques become more refined. Furthermore, the rapid evolution of generative models, such as the progression within the Qwen series, suggests that the predictive accuracy of these simulations will likely improve over time as base model capabilities increase. 
		
		It is important to note that the datasets utilized in this study—specifically the ELA dataset and BEA 2024 Shared Task sets—are relatively small compared to those used in other large-scale psychometric studies. While these sets provided sufficient data for an initial evaluation, larger and more diverse collections of student performance data may be necessary to fully validate the scaling laws and IRT parameter approximations observed here. A particularly compelling direction for future research involves the integration of model-generated rationales. While the current study focused on the final answer produced by the simulated student, the prompting architecture was designed to accommodate chain-of-thought processes. Engaging the ``thinking" component of the LLM to generate rationales could play a critical role in determining why certain items are difficult, potentially offering deeper diagnostic insights than result-based simulation alone. 
		
		Despite the encouraging Pearson correlations observed with larger models like Qwen-8B and Qwen-14B, the field is not yet at a point where LLMs can be reliably trusted for high-stakes difficulty prediction. The current models serve as useful tools for low-stakes automated pre-testing, but further rigorous testing and reliability assessments are required before they can replace live human pre-testing in formal assessment environments.
		
		\section*{Acknowledgments}
		
		The author gratefully acknowledges the Hawaii State Department of Education for their support of this research direction. The author would like to acknowledge Kai North, Frank Rijmen, and Suhwa Han for their discussions of the overall approach used in this paper. The author would like to thank Andrew Lan and Alex Scarlatos for their feedback on the draft and presentation of this work. 
		
		\bibliographystyle{plain}
		\bibliography{references}

		\appendix
		
		\section{Prompting}\label{sec:prompting}
		
		The model prompting consisted of two distinct parts; a system prompt that specified the way in which the model should behave, and the user prompt, that determined the task to be completed. We embedded the information regarding student abilities and their ordering in the system prompt, while the user prompt contained the question, the required student ability level, and the correct answer.
		
		\subsection{System Prompt}
		\begin{lstlisting}
			You are an AI model simulating a student's response to an assessment question. Your task is to generate a plausible answer, strictly adhering to the specified student ability level.
			
			**Ability Scale Context (Lowest to Highest):**
			[{', '.join(descriptors)}]
			
			* **Low-Ability Students (e.g., "Critical", "Deficient"):** Might get the answer correct by chance.
			* **Mid-Ability Students (e.g., "Basic", "Functional"):** More likely to give a correct answer.
			* **High-Ability Students (e.g., "Proficient", "Exemplary"):** Should provide the correct answer.
		\end{lstlisting}
		
		\subsection{User Prompt}
		\begin{lstlisting}
			This information is for your reference only, to understand the problem's solution.
			
			* **Question:**
			{question}
			
			* **Correct Answer:**
			{correct_answer}
			
			**Simulation Task**
			
			You must now generate a response from the perspective of a student at the following ability level:
			
			* **Student Ability Level:** {student_ability}
			
			Provide *only* the simulated student's response in the format below.
			
			**Student Answer:**
			[Your generated answer here]
		\end{lstlisting}

	\end{document}